\documentclass[letterpaper,twocolumn,10pt]{article}
\usepackage{usenix-2020-09}

\usepackage{tikz}
\usepackage{amsmath}

\usepackage[]{hyperref}
\usepackage{graphicx}

\usepackage{multicol}
\usepackage{amsfonts}

\usepackage[export]{adjustbox}
\usepackage{multirow}
\usepackage{times}
\usepackage{multicol, blindtext}
\usepackage[ruled,linesnumbered]{algorithm2e}
\usepackage{eqparbox,array}

\newcommand{\sm}{\textcolor{black}}

\newcommand{\name}{{eMoE}\xspace}
\usepackage{xcolor}
\usepackage{wrapfig}
\usepackage{tikz}
\usepackage{textcomp}
\usepackage{subcaption}

\newcommand{\DEL}[1]{\iffalse #1 \fi}

\newcommand{\squishlist}{
	\begin{list}{$\bullet$}
		{ \setlength{\itemsep}{0pt}
			\setlength{\parsep}{0pt}
			\setlength{\topsep}{0pt}
			\setlength{\partopsep}{0pt}
			\setlength{\leftmargin}{0em}
			\setlength{\labelwidth}{0em}
			\setlength{\labelsep}{0.2em} } }
	
	\newcommand{\squishlisttwo}{
		\begin{list}{$\bullet$}
			{ \setlength{\itemsep}{0pt}
				\setlength{\parsep}{0pt}
				\setlength{\topsep}{0pt}
				\setlength{\partopsep}{0pt}
				\setlength{\leftmargin}{2em}
				\setlength{\labelwidth}{1.5em}
				\setlength{\labelsep}{0.5em} } }
		
		\newcommand{\squishend}{
	\end{list}  }
\newtheorem{thm}{\textbf{Observation}}
\usepackage[most]{tcolorbox}
\definecolor{light-gray}{gray}{0.95}
\tcolorboxenvironment{thm}{
   center,
   boxsep=1.0pt,
   top=0.2pt,
   bottom=0.2pt,
    width=\linewidth,
    colframe=light-gray,
    colback=light-gray
}

\newcommand{\sh}[1]{\textcolor{blue}{[HS: #1]}}

\author{
{\rm Suraiya Tairin}$^{*}$, {\rm Shohaib Mahmud}$^{*}$, {\rm Haiying Shen}\\
University of Virginia
\and
{\rm Anand Iyer} \\Georgia Institute of Technology
}

\begin{document}

\title{eMoE: Task-aware Memory Efficient Mixture-of-Experts-Based (MoE) Model Inference}
\maketitle
\begingroup
\renewcommand\thefootnote{\hspace{-1mm}*}
\footnotetext{Equal contribution.}
\endgroup


\begin{abstract}

In recent years, Mixture-of-Experts (MoE) has emerged as an effective approach for enhancing the capacity of deep neural network (DNN) with sub-linear computational costs. However, storing all experts on GPUs incurs significant memory overhead, increasing the monetary cost of MoE-based inference. To address this, we propose eMoE, a memory-efficient inference system for MoE-based large language models (LLMs) by leveraging our observations from experiment measurements. eMoE reduces memory usage by predicting and loading only the required experts based on recurrent patterns in expert routing. To reduce loading latency while maintaining accuracy, as we found using the same experts for subsequent prompts has minimal impact on perplexity, \name invokes the expert predictor every few prompts rather than for each prompt. In addition, 
it skips predictions for tasks less sensitive to routing accuracy. Finally, it has task-aware scheduling to minimize inference latency by considering Service Level Objectives (SLOs), task-specific output lengths, and expert loading latencies. Experimental results show that compared to existing systems, eMoE reduces memory consumption by up to 80\% while maintaining accuracy and reduces inference latency by up to 17\%. It also enables processing prompts 40$\times$ longer, batches 4.5$\times$ larger, and achieves 1.5$\times$  higher throughput.

\end{abstract}

\maketitle 
\pagestyle{plain} 

\section{Introduction}
A recent trend in Deep learning (DL) has been increasing the model performance by increasing
the number of model parameters
~\cite{NEURIPS2020_1457c0d6, devlin2018bert}. 
However, the linear relationship between the computational cost and the model size remains as a crucial impediment to the practical applications of the large models.
To circumvent this bottleneck, sparse architectures~\cite{jacobs1991adaptive, shen2020q, dong2019hawq, dong2020hawq, child2019generating, lagunas-etal-2021-block} have been introduced. 
Among these, the Mixture-of-Experts (MoE) stands out as one of the most prominent sparse models. Its adoption has substantially amplified the scalability of DNN models across various DL applications, encompassing natural language processing~\cite{DBLP:journals/corr/abs-2112-06905, fedus2022switch, dai2024deepseekmoe}, computer vision~\cite{MLSYS2023_5616d34c}, speech recognition~\cite{you21_interspeech, you20223m}, and recommendation systems~\cite{ma2018modeling}.

An MoE layer comprises a gate and a collection of experts. The gate selects only a small number (e.g., 1 or 2)  of experts for computation based on each input. 
Its sparse expert activation enables considerable expansion of model size by several orders of magnitude without a commensurate increase in computational requirements (FLOPs). However, the scaling of such models demands an excessively large amount of scarce GPU memory~\cite{kwon2023efficient}. This directly corresponds to substantial monetary expenses when implementing MoE-based inference systems at scale. Our data analysis, detailed in \S \ref{sec:motivation}, reveals that an MoE-based model typically consumes 4x to 14x more memory compared to its dense counterpart. Due to the intricate dependence between model capacity and memory requirements, designing a memory-optimized MoE inference system is crucial for serving very large models. 
\DEL{Additionally, MoE-based models exhibit task sensitivity, where certain tasks are more influenced by expert selection than others, further complicating the design of an efficient inference system.}
Existing state-of-the-art inference systems such as vLLM~\cite{kwon2023efficient}, DeepSpeed-FastGEN~\cite{dsfgen}, Sarathi-Serve~\cite{agrawal2024taming}, DistServe~\cite{zhong2024distserve} are designed for the traditional transformer architecture and thus unequipped to provide memory-efficient MoE inference. 


In this work, we propose eMoE, a memory-efficient inference system for MoE-based  large language models (LLMs). 
\name is designed by leveraging our observations from experiment measurements. \name integrates several components working in unison:

\squishlist
\item \textbf{Expert Prediction.} We observe a recurring pattern in token-to-expert routing, where certain experts process more tokens than others within a given time period (i.e., popular experts). Based on this observation, we propose an on-demand loading of popular experts during inference to reduce memory footprint. By leveraging the prior distribution of expert selection, \name predicts and proactively loads the experts for future inputs.

\item \textbf{Periodic Expert Invocation.} 
However, we found that time-efficient on-demand expert loading is challenging due to the increased inference latency. We observe that subsequent prompts are highly correlated, and using the same experts for subsequent prompts does not significantly affect perplexity. To reduce expert loading overhead while maintaining accuracy, \name invokes the expert predictor every few prompts rather
than for each prompt.

\item \textbf{Task-aware Expert Loading.} 
Additionally, our data analysis reveals that an MoE-based model can produce the same output with inaccurate experts for certain tasks as it does with the correct experts. Leveraging this observation, \name skips predictions for
tasks less sensitive to token-to-expert-routing accuracy.

\item \textbf{Task-aware Request Scheduling.} Furthermore, leveraging the above observation with another observation that certain tasks generate significantly fewer tokens than others, \name uses profiled task-specific token generation length, task-aware expert loading latency and user-imposed inference latency Service Level Objective (SLO) in request scheduling  with an objective to minimal impact on the end-to-end inference latency. It offers an advantage over existing request schedulers~\cite{yu2022orca,kwon2023efficient,agrawal2024taming,zhong2024distserve} by considering expert loading latency.

\squishend

\DEL{Existing state-of-the-art request scheduling systems for LLM inference such Orca~\cite{yu2022orca} and vLLM~\cite{kwon2023efficient} which only consider available resources in scheduling decision cannot work optimally for \name. Furthermore, these systems do not leverage the task-specific generation length characteristic. This suggests the existing request scheduler may not be sufficiently effective in \name. To address the limitations of these systems, }



Unlike previous methods that rely on continuous expert prefetching during inference~\cite{hwang2024pre,xue2024moe,yi2023edgemoe}, \name periodically loads and reuses experts, reducing the inference latency associated with these approaches. The limitations of these methods are discussed in \S\ref{sec:existing}. Our real experimental evaluations show that compared to existing systems, eMoE reduces memory consumption by up to 80\% while maintaining accuracy and reduces inference latency by up to 17\%. It also enables processing prompts 40$\times$ longer, batches 4.5$\times$ larger, and achieves 1.5$\times$  higher throughput. 


In summary, we make the following contributions: 
\squishlist

 \item We systematically analyze token-to-expert recurrence patterns and task-specific characteristics, such as tolerance for less accurate experts and varying output lengths.  Our findings reveal that existing solutions for memory-efficient MoE inference are insufficient in addressing both memory optimization and inference latency effectively.
    \item We propose eMoE, a memory-efficient inference system that uses expert prediction to load experts, incorporates periodic expert invocation, employs task-aware expert loading to minimize latency for less routing-sensitive tasks, and utilizes task-aware request scheduling to optimize both inference latency and memory consumption. 

    \item We evaluate eMoE against state-of-the-art transformer-based inference systems, demonstrating significant reductions in memory consumption while preserving accuracy and minimizing end-to-end inference latency.
    
\squishend
\vspace{-0.1in}
\section{Observation and Motivation}
\label{sec:motivation}

{\subsection{Experiment Settings}
\label{sec:analysis_sett}
We conducted an analysis experiment on four popular publicly available MoE based LLMs: Mixtral-8x7B \cite{jiang2024mixtral}, Mixtral-8x22B, TinyMixtral \cite{jiang2024mixtral} and OpenMoE \cite{xue2024openmoe}. 
We used various popular natural language processing (NLP) tasks including
text summarization, question answering (QA), 
semantic similarity comparison and semantic classification. 
\begin{table}[t] 
    \centering
    \footnotesize
    \caption{Dataset summary. \label{table:data_tasks}}
    \begin{tabular}{|p{0.12\textwidth}|p{0.14\textwidth}|p{0.16\textwidth}|}
    \hline
    Dataset & Contents & Task \\
    \hline
    XSUM~\cite{zhang2020pegasus} & News articles & Summarize (SUM)\\
    \hline
    Tweet-sentiment~\cite{tweet-finance} & Tweets related to financial markets, stocks, and economic discussions & Classify tweet sentiment (CLSFY)\\
    \hline
    SQUAD~\cite{squad} & Context and question pairs & Question answer (QA)\\
    \hline
    Headlines similarity~\cite{silcock2024massive} & Pairs of headlines on same topics & Compare semantic similarity (COMP)\\
    \hline
    Human-assistant conversation \cite{converse-data} & Pairs of prompts and responses & Conversation (CONV)\\
    \hline
    \end{tabular}
\end{table}
Details of the datasets used for these tasks are summarized in Table \ref{table:data_tasks}. Following \cite{yu2022orca} and \cite{kwon2023efficient}, we generated synthetic client request traces since there is no publicly available request trace for  generative MoE-based language models. The generated trace follows the Poisson distribution. To generate a set of requests, first, we drew the number of requests from the Poisson distribution. Next, for each arrival of requests, we generate tasks following a multinomial distribution, where tasks have a uniform probability. 
We conducted our experiment on a computer equipped with Intel Xeon processor with 128GB host memory and four Nvidia A100 Tensor core GPU with 40GB device memory. We set the maximum number of generated tokens to 1000 for all the models to avoid out-of-memory errors. For accuracy measurement, we consider the output of an MoE-based model with all experts as the ground truth. We measured the semantic similarity between an output and ground truth as the accuracy using a pre-trained BERT model~\cite{devlin2018bert}.

\subsubsection{High Inference Latency with Dynamic Expert Loading During Inference} To reduce the memory requirements of MoE models, 
an alternative strategy involves keeping all experts on the CPU and dynamically transferring the router-selected experts to GPUs during inference, instead of pre-loading all experts onto the GPU. \DEL{In this analysis, we keep all the experts in the CPU, and when an input request is received, the necessary experts are transferred onto the GPUs for each layer, following the router's expert selection. When an input is received, it undergoes sequential processing through the model layer by layer. Specifically, for each MoE layer, the router selects experts, and the input tokens are routed to those chosen experts. Therefore, to reduce the memory consumption only the required experts can be transferred instantly into GPU memory, preventing the need to load all experts, which could incur substantial memory consumption.} However, during inference, the communication between the CPU and GPUs for transferring the required experts introduces additional time overhead. 
To assess this trade-off, we conducted experiments comparing the inference time for the two cases.
Figure \ref{fig:dynamic}(a) shows that 
the inference time is significantly higher in dynamic loading than in static pre-loading. It is 3.2x higher for OpenMoE, and 5x higher for Mixtral-8x7B.  
Figure \ref{fig:dynamic}(b) exhibits the corresponding memory consumption for each case. The memory consumption of dynamic loading is lower than the static pre-loading. It is  1.5x lower for OpenMoE, and 1.4x lower for Mixtral.  The average number of active experts is around 67\% and 78\% for OpenMoE and Mixtral respectively.
Further analysis reveals that the average time to transfer experts from the CPU to the GPUs is $\sim$4.431 seconds and $\sim$12.744 seconds for OpenMoE and Mixtral, which is 2.2x and 3.6x of the inference time of OpenMoE and Mixtral, respectively. 
These findings highlight a critical trade-off: while dynamic loading reduces memory consumption, it significantly increases inference latency. As a result, instant loading of experts during inference is not a practical solution for memory-efficient MoE inference.

\begin{figure}[!htb]
\vspace{-0.15in}
    \centering
   {\includegraphics[height=3cm, width=8.5cm]{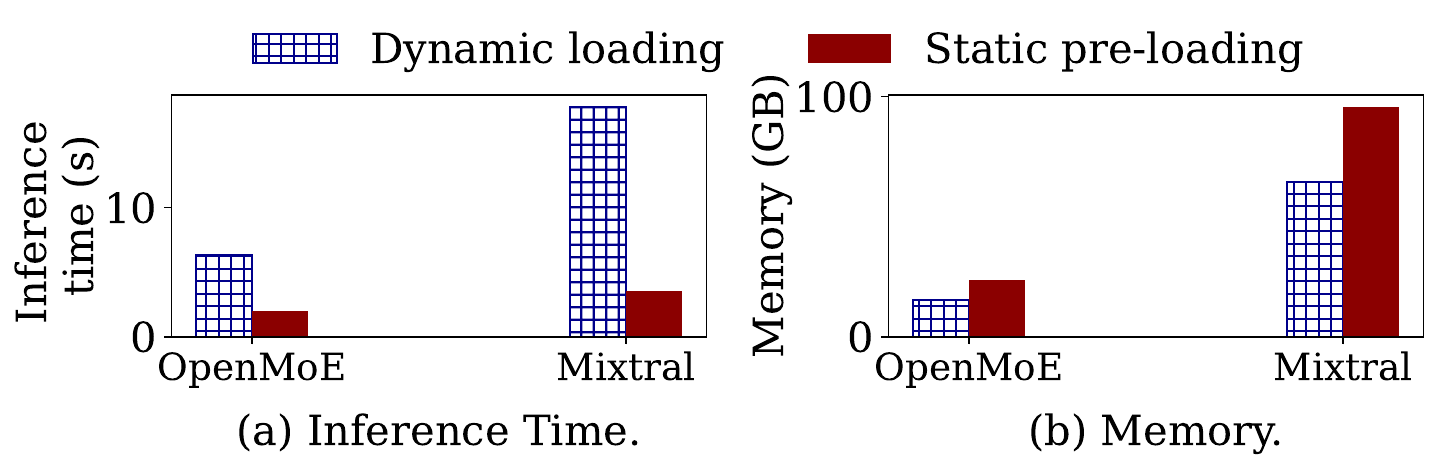}}
   \caption{Inference time with memory consumption for dynamic expert loading during inference.}
   \label{fig:dynamic}  
   \vspace{-0.1in}
\end{figure}


\begin{thm}\label{Observation 3} Instant loading
of the required experts during inference is not a feasible solution to reduce the memory consumption of MoE models as it significantly increases inference latency.
\end{thm}

\begin{figure}[!htb]
   \vspace{-0.15in}
    \centering
   {\includegraphics[height=3cm, width=8.5cm]{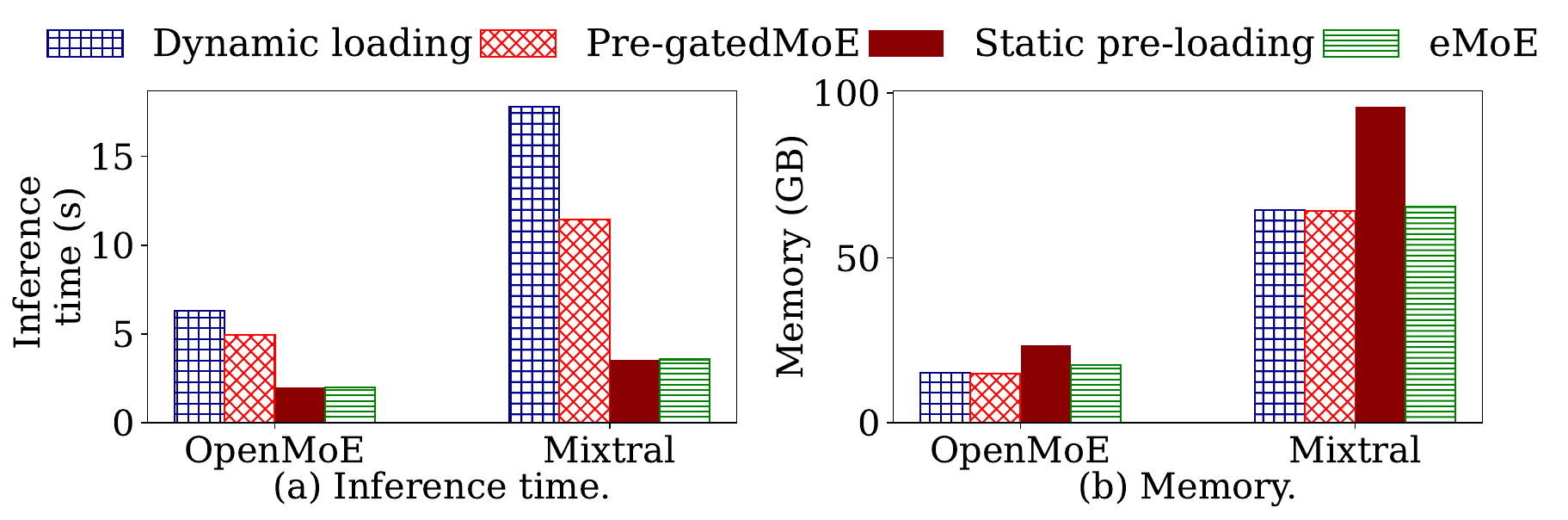}}
   \vspace{-0.1in}
   \caption{Inference time with memory consumption for different approaches.}
   \vspace{-0.15in}
    \label{fig:pregate}  
\end{figure}

\subsection{Issues with Existing Approaches}
\label{sec:existing}
Some existing approaches, like Pre-gatedMoE \cite{hwang2024pre}, MoE-Infinity \cite{xue2024moe}, and EdgeMoE \cite{yi2023edgemoe}, reduce communication overhead by overlapping computation and data transfer, prefetching experts for the next layer during the current layer's computation into the GPU. To assess the impact of these techniques, we evaluated Pre-gatedMoE against dynamic loading, static pre-loading (all experts on GPU), and our proposed eMoE, which predicts, periodically loads, and reuses experts. Figure \ref{fig:pregate} shows that Pre-gatedMoE reduces latency by 1.2x–1.6x compared to dynamic loading but has 2.5x–3.5x higher latency than static pre-loading and eMoE. Real-time expert transfer between CPU and GPU can lead to memory and bandwidth contention, especially when the GPU is concurrently handling computations. This contention impacts both data transfer and processing speed, resulting in increased latency in Pre-gatedMoE. Additionally, it requires retraining the MoE model with its gating mechanism, which is resource-intensive and time-consuming.

MoEInfinity and EdgeMoE both utilize expert prefetching to reduce memory usage but introduce complexity and overhead. MoEInfinity employs "group activation" to prefetch frequently co-activated experts but requires maintaining an expert activation matrix for each inference request, adding computational overhead. EdgeMoE is tailored for edge devices with limited GPU memory, making it less suitable for larger models or multi-GPU setups. Both methods increase inference latency and require either full MoE model training or creating additional data structures, complicating their use with larger models.



\begin{figure}[!htb]
\vspace{-0.15in}
    \centering
    \subfloat[Summarization. \label{fig:p1}]{\includegraphics[height=3cm, width=3.8cm]{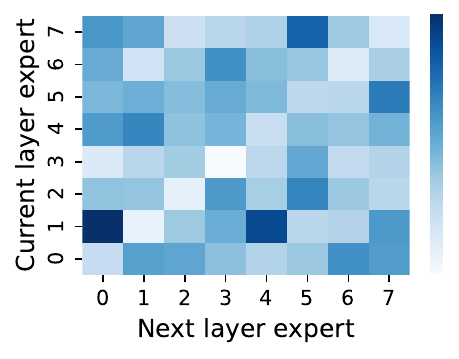}}
    \hspace{-0.09in}
    \subfloat[Question-answering. \label{fig:p2}]{\includegraphics[height=3cm, width=3.8cm]{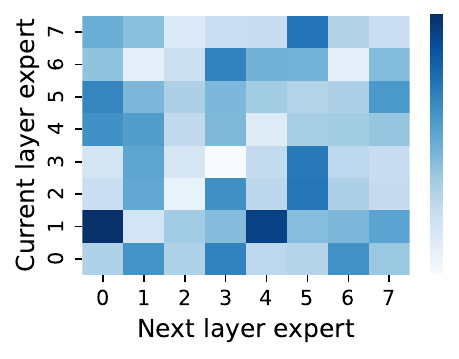}}
    \vspace{-0.1in}
   \caption{Expert activation transition patterns between consecutive layers.}
   \vspace{-0.15in}
    \label{fig:pattern}   
\end{figure}

\vspace{-0.15in}
\subsubsection{Repetitive Patterns in Expert Activation}
To investigate the repetitive patterns in expert activation between consecutive layers in MoE model, we conducted experiments to track expert selection across layers. For each input, we recorded the pair of experts chosen at consecutive layers, compiling this data into a matrix where each entry at the position 
$(i,j)$ represents the number of times Expert 
$i$ at the current layer was followed by Expert 
$j$ at the next layer.
We use a heatmap to visualize these frequencies. Figure \ref{fig:pattern} presents the heatmap for summarization and question-answering tasks for Mixtral-8x7B. In the figure, darker colors (e.g., deep blue) indicate a higher frequency of transitions, while lighter colors (e.g., pale blue or white) indicate lower frequencies. 
The figure reveals strong, consistent patterns in expert activation between layers, suggesting an opportunity to predict future expert activations based on prior patterns. This insight highlights the potential for developing a predictive method to anticipate expert selection, which could significantly reduce memory usage.\looseness=-1

\DEL{case 1: when the router selected experts are transferred dynamically onto GPU during inference, and case 2: when all the experts are pre-loaded in the GPU before any inference starts and there is no transfer of experts during inference. When all the experts are loaded to GPUs, it will take a huge memory, increasing the number of GPUs needed for such huge memory. {\sh{indicate how many GPUs are needed in case 1 and case 2}}-indicated in the below description- done}

\vspace{-0.1in}
\subsubsection{Inference Latency is Task Sensitive}
\label{shorcomings}
The inference latency is directly proportional to the number of generated tokens. Therefore, the inference latency may vary depending on the task's characteristics. To test our hypothesis, we analyze a number of 
popular NLP tasks given in Table~\ref{table:data_tasks}.

\begin{figure}[!t]
\centering
\subfloat[OpenMoE\label{fig:OpenMoE_task_lat}]  {\includegraphics[width=4cm, height=4cm]{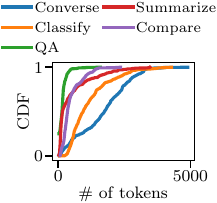}}
\hspace{0.05in}
\subfloat[Mixtral\label{fig:mixtral_task_lat}]{\includegraphics[width=4cm,height=4cm]{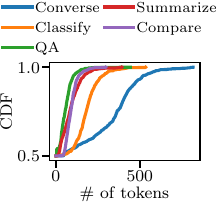}}
\vspace{-0.1in}
\caption{CDF of the number of generated tokens across different tasks}
\vspace{-0.15in}
\label{fig:task_len}
\end{figure}
Figure~\ref{fig:task_len} shows the Cumulative Distribution Frequency (CDF) of requests versus the number of generated tokens. The figure reveals variations in the lengths of generated outputs across different tasks. Specifically, the QA tasks yield the shortest output length, while the conversation tasks produce the longest output length. Notably, both models generate explanations \sm{in their generated texts} for classification and comparison tasks, resulting in longer output lengths compared to the QA tasks. Additionally, the 90th percentile of the output length of the QA tasks is 17\% and 26\% smaller than that of the conversation tasks for the OpenMoE and Mixtral models, respectively. The rest of the tasks' 90th percentile output lengths follow this order: QA (Mixtral: $2.2$ sec, OpenMoE: $1.8$ sec) $>$ Summarization (Mixtral: $1.8$ sec, OpenMoE: $1.4$ sec) $>$ Semantic similarity measure (Mixtral: $1.4$ sec, OpenMoE: $1.1$ sec). The output length difference among the tasks is caused by their output characteristics. 
Therefore, we make the following observation
\begin{thm}
    \label{obs:latency}
    Different tasks exhibit different output length distributions. The length difference can be often as high as twice the output length of a certain task. 
\end{thm}

\vspace{-0.2in}

\subsubsection{Tasks Show Varying Sensitivity To Tokens-to-expert Routing Accuracy}
\label{sec:task-moe-sen}
Existing work indicates that layers closer to the input in neural networks tend to learn general representations, while layers closer to the output specialize in task-specific representations~\cite{zeiler2014visualizing, rajbhandari2022deepspeed}. Building upon this observation, we hypothesize that certain tasks may not require highly accurate representations in the layers closer to the input. This suggests that the inference system can prioritize experts selections for tasks that demand more accurate representation. 

\DEL{To understand it, we vary the number of MoE layers for which a token is routed to its top experts, and route a token to some random experts in the other MoE layers. Starting with the first MoE layers, we progressively make a certain percent of MoE layers inaccurate\sh{how?} and record the model outputs.} 

\begin{figure}[!t]
    \centering
    \includegraphics[width=8.5cm, height=3.5cm]{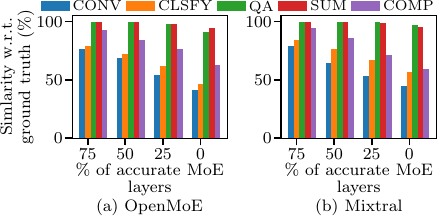}
    \caption{Accuracy with progressive applying of exact token-to-expert routing in MoE layers closer to output layer.}
    \label{fig:acc_analysis}
\end{figure}

To ascertain the hypothesis, we conducted an analysis experiment. In this experiment, starting from the first MoE layer, we progressively made the MoE layers' routing inaccurate and measured the output accuracy. 
The results of this analysis experiment are shown in Figure~\ref{fig:acc_analysis}, where the X-axis denotes the percentage of MoE layers where accurate token-to-expert routing was applied and Y-axis represents the similarity. 
We see that for the classification and compare tasks, both models maintain above 90\% similarity with respect to the full model even when all tokens are routed incorrectly. We also note that open-ended tasks such as conversation and summarization show low tolerance to inaccurate expert routing. Even with 75\% accurate MoE layers, the similarity score falls below 80\%. On the other hand, for the QA tasks which need to be more specific as the answer is in the context, the similarity score is above 80\% for both models when only 50\% of the MoE layers have accurate routing. For the summarization tasks, the accuracy falls below 80\% when 50\% of the MoE layers are accurate. We see that a task's tolerance to MoE layers' routing accuracy varies greatly. Therefore, we make the following observation:

\begin{thm}
    \label{obs:slo_acc}
    Different tasks exhibit different tolerance to inaccurate MoE layers' expert routing. Tasks that are more open-ended such as conversation demand more accurate routing than tasks that are more specific such as semantic classification.
\end{thm}

\vspace{-0.2in}
\section{System Design of eMoE}
\vspace{-0.1in}
\subsection{Overview}



\begin{figure}[!htb]
\vspace{-0.2in}
    \centering
    \includegraphics[width=0.46\textwidth]{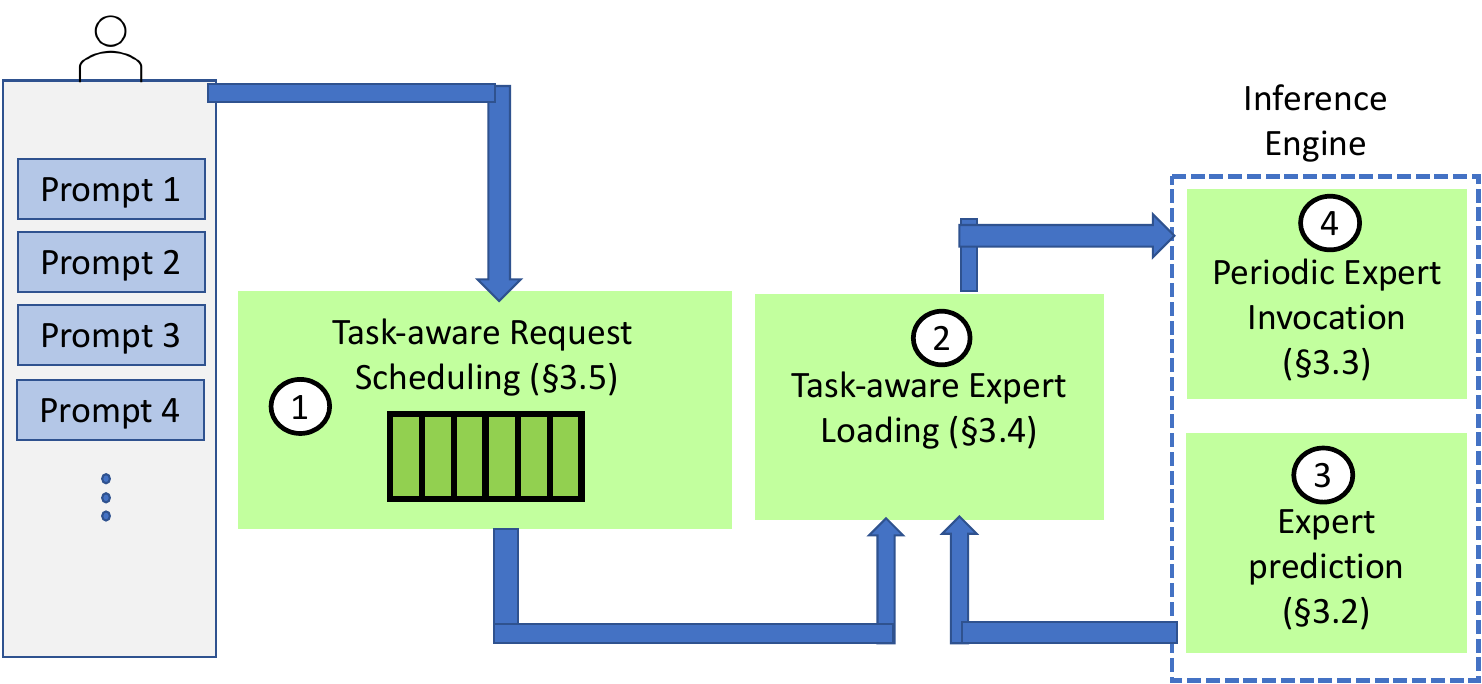} 
    \vspace{-0.15in}
    \caption{Overview of eMoE.}
    \vspace{-0.15in}
   \label{fig:design}    
\end{figure}

Figure \ref{fig:design} illustrates the architecture of the \name inference system. The inference process starts with task type extraction. After extracting the task type, the prompts are scheduled by eMoE's task-aware request scheduler (\textcircled{1}). 
The scheduler aims to minimize the inference latency by considering profiled task-specific token generation length, user-imposed latency SLO and expert loading latency. The scheduled requests are sent to the task-aware expert loading module (\textcircled{2}). This module leverages tasks' sensitivity to token-to-expert routing accuracy and the expert prediction to selectively load experts to the inference engine to minimize expert loading latency. In addition to serving inference requests, the inference engine also runs the expert prediction model  based on recurrence
patterns (\textcircled{3}). To save the overhead for expert loading while maintaining accuracy, it periodically invokes expert prediction at regular intervals (\textcircled{4}). 
Once expert loading is initiated, the inference engine starts serving the scheduled requests. In the following, we present each of these components.

\vspace{-0.15in}
\subsection{Expert Prediction}
In the inference pipeline, we first identify the task type for each request for task-aware expert loading. Next, we predict the specific experts needed and load only predicted experts onto the GPU. We present task-type extraction and expert prediction methods in \S \ref{sec:task-type-extract} and \S \ref{sec:prediction}. The MoE inference process using the expert predictor is described in \S \ref{sec:inference_predictor}.

\vspace{-0.1in}
\subsubsection{Task Type Extraction}
\label{sec:task-type-extract}
This task type extraction method identifies the task type associated with each individual user request. \sm{The method leverages the occurrence of certain task-specific keywords to determine a request's task type. A task is typically described by one or two sentences and task description for a task type is likely to have similar keywords. For example, a user may write ``Summarize the following text'' or ``Write a synopsis for the following text'' to request for a summarizing task. 
} 
Upon receiving a user request, this method processes the input tokens and generates a potential task type as the output. Specifically, it searches for some profiled keywords in the sentences in the input. \sm{Each identified keyword is compared against the keywords from all task types. The task type that has the highest number of matches is assigned as the task type of the input.} The method runs on an on-demand basis on CPU and therefore does not interfere with the inference system pipeline. \looseness = -1

\subsubsection{Task-aware Expert Prediction}
\label{sec:prediction}
Observation \ref{Observation 3}  shows that instantly loading the required experts during inference in MoE models to reduce memory consumption increases inference latency. Using fewer experts than the model was trained with reduces memory consumption but compromises output quality \cite{rajbhandari2022deepspeed}. Besides, the workload among experts varies dynamically during inference \cite{li2023accelerating, MLSYS2023_5616d34c}. An ideal solution should reduce memory consumption without harming the quality of model output or increasing inference latency. The most effective approach is to proactively determine the necessary experts for an inference request before processing it and load only those experts into the GPU's memory. Building on these insights, we propose an expert prediction method to reduce memory consumption while maintaining model output quality and inference latency.


We employ a machine learning (ML) based method to predict the experts. This predictor forecasts the future experts needed for an incoming request based on the prior distribution of experts. 
The experts are represented by numerical digits. For instance, if there are $m$ MoE layers in an MoE model, and if we denote the expert sequence for layer $i$ as $e_i$, the sequence of the expert index for $m$ layers can be represented as follows: $e_1, e_2,...e_i,..., e_m$. Each $e_i$ represents a series of $k$ indices for top-k routing. 
This expert prediction can be seen as a sequence prediction task, where elements in the sequence exhibit certain correlations or patterns. Typically, future elements in a sequence depend on the past elements. In sequence or series prediction, the output is a distribution of probabilities over possible outcomes, representing the relative frequency of each possible next element, rather than a single fixed value. The element with the highest probability is considered as the next element in the sequence. 

\DEL{Experts in MoE models are trained to specialize in different types of input, and a token's expert selection demonstrates a pattern across the MoE layers. }

\textbf{Correlation in Expert Sequence.} We analyze expert selection correlation and use cross-correlation \cite{cross-co} to measure the correlation between experts chosen in one layer and those chosen in the next layer. For a top-k routing, 
we consider the $k$ chosen experts in layer $i$ as a sequence $e_i$ and the $k$ chosen experts in layer $i+1$ as a sequence $e_{i+1}$, and compute the correlation between them. 
We compute the average cross-correlation of selected experts from one layer to the next for 1000 prompts using the settings in \S \ref{sec:analysis_sett}. Figures \ref{fig:cross_om} and \ref{fig:cross_mx} show the correlation for each layer. 
For both OpenMoE and Mixtral-8x7B, we observe a significant correlation of around 0.50 between the experts selected by consecutive layers. This is because the gate network uses basic features such as parts of speech and word meaning to select experts, resulting in similar tokens being handled by similar experts across layers \cite{li2023accelerating}. This consistent selection process leads to a high degree of correlation between the experts chosen in consecutive layers. We leverage this correlation to estimate the expert in the next layer using an ML model.


\begin{figure}[!htb]
    \centering
    \subfloat[OpenMoE.\label{fig:cross_om}]{\includegraphics[height=2.5cm,width=4.2cm]{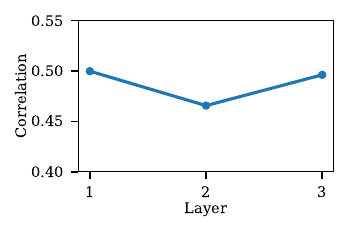}}
    \hspace{-0.14in}
    \subfloat[Mixtral.\label{fig:cross_mx}]{\includegraphics[height=2.5cm,width=4.2cm]{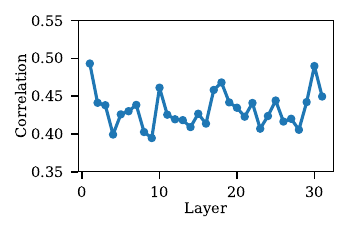}}
   \vspace{-0.12in}
    \caption{Correlation in expert sequence between consecutive layers.}
    \vspace{-0.16in}
   \label{fig:analysis_cross}    
\end{figure}



We then calculate the cross-correlation to examine the relationship between experts selected for consecutive prompts. Figures \ref{fig:cross_p_om} and \ref{fig:cross_p_mx} show the correlation coefficient between each prompt and the following prompt. In OpenMoE and Mixtral-8x7B, the correlations range from 0.4 to 0.6 and from 0.75 to 0.95, respectively, indicating a strong correlation in expert selection between prompts. There exists shared context or vocabulary 
across consecutive prompts. This shared information results in a high correlation in expert selection between consecutive prompts.

\begin{figure}[!htb]
    \centering
    \subfloat[OpenMoE.\label{fig:cross_p_om}]{\includegraphics[height=2.5cm,width=4.2cm]{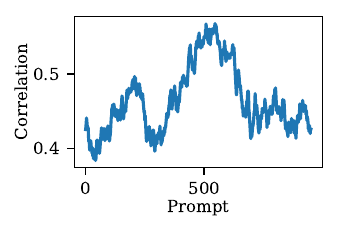}}
    \hspace{-0.11in}
    \subfloat[Mixtral.\label{fig:cross_p_mx}]{\includegraphics[height=2.5cm,width=4.2cm]{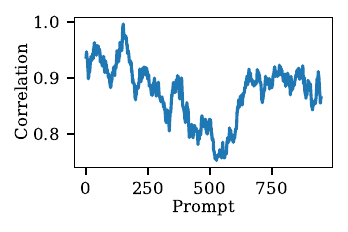} }
    \vspace{-0.1in}
    \caption{Correlation in expert sequence between consecutive prompts.}
    \vspace{-0.2in}
   \label{fig:analysis_cross_p}    
\end{figure}


\textbf{Preparing Expert Sequence for Prediction.}
\label{sec:prepare_predictor}
Motivated by the above observations, we predict the future expert sequence based on past expert sequences. 
For an incoming prompt, before processing it, we must determine the expert sequence (i.e., the experts in each layer) which is the future expert sequence for all its tokens. For this future expert sequence, the past expert sequence would be the expert sequence of previous layers. 
We consider two cases to predict the expert sequence: 
(1) predicting future expert sequence for one layer at a time using the past layer's expert sequence (layer-by-layer prediction), denoted as eMoE-L, and (2) predicting future expert sequence for all the layers at once using the past prompt's expert distribution (all-layer prediction), denoted as eMoE-A. 
Suppose there is an incoming inference request $r_1$. 
For case (1), let the predicted expert of the $i$-th layer for prompt $r_1$ be denoted as $e_{i}^{r_1}$. Then the prediction method $f$ can be represented as follows: $e_{i}^{r_1} = f(e_{i-1}^{r_1})$.
To predict the experts for the first layer, we use the experts of the first layer of the previous prompt. 

For case (2), suppose there are two inference requests $r_1$ and $r_2$.
We want to predict the expert sequence of $r_2$ using the expert selection of $r_1$. The experts selected by the router for all the layers for $r_1$ are denoted as $e_1^{r_1}, e_2^{r_1},...,e_m^{r_1}$, where the model has $m$ layers. Let's assume the predicted expert sequence for all layers of $r_2$ is $e_1^{r_2}, e_2^{r_2},...,e_{m}^{r_2}$. Then, the prediction method $f$ can be presented as follows: \vspace{-0.05in}
\begin{equation}
e_1^{r_2}, e_2^{r_2},...,e_{m}^{r_2} = f(e_1^{r_1}, e_2^{r_1},...,e_m^{r_1}).\vspace{-0.05in}
\end{equation}

\DEL{where the prediction method takes the previous prompt $r_1$'s expert sequence $e_1^{r_1}, e_2^{r_1},...,e_m^{r_1}$ as input and outputs the expert sequence $e_1^{r_2}, e_2^{r_2},...,e_{m}^{r_2}$ for $r_2$.
{\sh{very confusing. you want to use experts of previous layers to predict subsequent layers, but "previous layers" are for the same request. Also, how to predict the first layer's expert?}}-done 
}

\DEL{{\color{red}We consider both of these cases in our approach. Because the input and output differ between the two, each case requires its own predictor. We design separate predictors for each case, treating them as distinct approaches. }}
}


\textbf{Expert Predictor Selection.} 
\label{sec:select}
Selecting a suitable ML model for sequence prediction is crucial for effectively learning relationships between past and future sequences. We choose a transformer-based model given its ability to capture and maintain dependencies in very long sequences using the attention mechanism \cite{vaswani2017attention}.
In contrast, other models for sequence prediction, such as LSTM \cite{hochreiter1997long}, struggle to capture dependencies in long sequences that are far apart, as they process sequences sequentially and suffer from vanishing gradient problems in very long sequences \cite{pascanu2013difficulty}. Transformer-based models often perform comparably or better than traditional sequence prediction models \cite{gruver2023large}. 
We use the GPU to run our expert predictor to minimize latency, as it consumes very little GPU memory. Our empirical results in \S \ref{sec:predictor_performance} show that the predictor typically uses only 0.24\%-1.3\% of the MoE model's memory.
When loading predicted experts, we compare them with those already loaded in the GPU. New experts are identified and loaded into the GPU, while experts not in the prediction are moved to the CPU. Experts are loaded according to a memory budget, prioritizing those with the higher workload, which is determined by the total number of tokens they receive. If the expert selected for a token is not on the GPU, the token is routed to the next top-k expert that is on the GPU.
\vspace{-0.1in}
\subsection{Periodic Expert Invocation}
\label{sec:inference_predictor}

\DEL{To reduce the memory consumption, the expert predictor should be used for MoE inference system efficiently. The basic approach would be calling this expert predictor for each prompt. We can call this expert predictor for each prompt before processing it. The expert predictor will predict experts for this prompt. We can load only the predicted expert onto the GPU memory so that all the experts will not be loaded. In this way, the GPU memory consumption will be reduced. Then the MoE model will process the prompt using the predicted expert which is loaded onto the GPU.  However, this basic approach introduces several problems in the system. {\color{red} First, the invocation of the predictor and loading the experts of the basic approach} would introduce certain delays which increase the inference time.{\sh{previous half sentence and the following half sentence are the same?}}-done   
}

\DEL{{\color{red}Second, if the expert predictor is placed on the GPU, it would consume GPU memory. However, as demonstrated in Section \S \ref{performance_prediction} and \S \ref{effect_loading}, our empirical findings indicate that the expert predictor uses only a negligible amount of memory, typically between 0.01\% and 0.04\% of the MoE model's total memory. Additionally, loading the predicted experts reduces the overall memory consumption by a significant 41\% to 42\%. 
Therefore, a key challenge here is how to use this expert predictor without causing performance degradation in the main MoE inference system, ensuring that inference times are not delayed}{\sh{didference between "affect" and "delay" here?}} -done. {\sh{people will ask using an extra LLM will consume more memory resource, how come this memory consumption will be less than the memory saved?}}-done}

Calling the expert predictor for each inference request adds latency to the inference time (shown in \S \ref{sec:ex_overhead}), which is undesirable. Therefore, calling the predictor should be strategized to avoid harming the inference latency performance of the MoE inference system. This means the expert predictor should be called opportunistically, not for every inference request. One straightforward approach is to invoke the expert predictor at regular or adaptive intervals, rather than for every inference request. The predicted experts can then be used for several consecutive requests without needing to re-invoke the expert predictor for each one. 
We assume that there is a level of stickiness or continuity among subsequent incoming inference requests. 
For example, in a conversation, there is often a logical flow or sequence to the topics being discussed. Therefore, subsequent inference requests are often related to the previous ones. 

\DEL{{\sh{the rationale is not clear. does it mean some requests won't be predicted? why these requests do not need to predict. In the design, you always save something but do not compromise something else.}-done}}

\begin{wrapfigure}{c}{4.5cm}
\vspace{-0.3in}
    \centering    \includegraphics[width=0.25\textwidth, height=2.7cm]{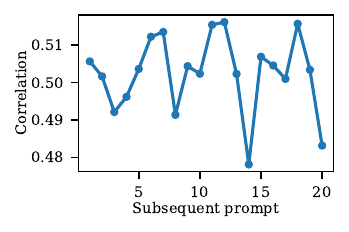} 
    \vspace{-0.25in}
    \caption{Correlation in consecutive prompts.}
    \vspace{-0.1in}
   \label{fig:subsequent}    
\end{wrapfigure}



To observe the correlation between subsequent prompts, we conduct an experiment using the settings in \S \ref{sec:analysis_sett}. For each prompt, we compare its correlation \cite{cross-co} with the subsequent prompt. Specifically, for a prompt, we consider pairs with the $i$-th subsequent prompt ($i=1,2,…,20$) and measure the correlation of each pair. 
Figure \ref{fig:subsequent} illustrates the average correlation between a prompt and each subsequent prompt. We observe that the correlation is significantly high, ranging from 0.48 to 0.55. This is due to the shared context or common vocabulary. When prompts are contextually related or use similar terms, their contents become more similar, leading to higher correlation. Words and phrases that frequently appear together or within the same context 
results in stronger correlations.

\begin{figure}[!htb]
\vspace{-0.05in}
    \centering
    \subfloat[OpenMoE.\label{fig:same_om}]{\includegraphics[height=2.7cm,width=4.4cm]{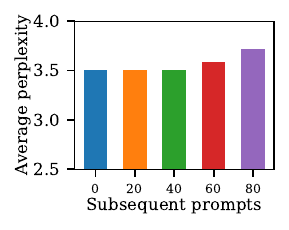} }
    \hspace{-0.2in}
    \subfloat[Mixtral.\label{fig:same_mx}]{\includegraphics[height=2.7cm,width=4.2cm]{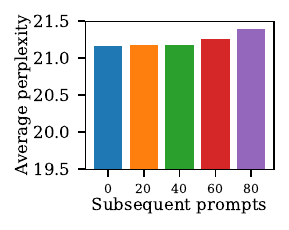}}
    \vspace{-0.1in}
    \caption{Perplexity using same experts.}
    \vspace{-0.15in}
   \label{fig:analysis_same}    
\end{figure}

To investigate whether experts chosen for one prompt can be reused for subsequent prompts, we conducted an experiment using the settings described in \S \ref{sec:analysis_sett}. In this experiment, we used the same experts selected for a prompt for its next \(p\) prompts, with \(p\) values of 0, 20, 40, 60, and 80. \(p = 0\) means the same experts are not reused for any subsequent prompts. Figure \ref{fig:analysis_same} shows the average perplexity for different numbers of subsequent prompts. We observed that the average perplexity remains nearly the same for 20 and 40 prompts but starts increasing from 60 onwards.
This is because the prompts are highly correlated, and using the same experts for some subsequent prompts does not significantly impact perplexity. 

Motivated by the above observations, we consider that the expert predictor does not need to be invoked for every inference request. Once the expert predictor is called and the predicted experts are loaded, they can be reused for multiple subsequent inference requests in the future. Therefore, in our design, we consider calling the expert predictor at regular intervals (i.e., after a certain number of prompts). \DEL{\color{red}Specifically, we consider invoking the expert predictor after a certain number of prompts. The overview of our
inference system with expert prediction is demonstrated in Figure \ref{fig:overview} 
}
\DEL{Suppose the number of prompts is set to $i$ prompts for the system. This means that after every $i$ prompt, our system calls the expert predictor. 
}
In our inference system, we maintain an index of the incoming requests as 0, 1, 2, and so on. Based on this index, we invoke the expert predictor after every $p$ prompts. For each incoming request, the system first checks if it is the $p$-th prompt. If not, the system processes the prompt using the experts already loaded on the GPU without invoking the predictor. Otherwise, the system calls the expert predictor, loads the predicted experts onto the GPU, and then processes the prompt using these newly loaded experts. For the first prompt, the system loads all the experts onto the GPU, and subsequently, after every $p$ prompts, it invokes the predictor and loads only the predicted experts. 
\DEL{In this way, only the predicted experts by the expert predictor are loaded into the GPU memory which significantly reduces the GPU memory consumption during inference.} 


\subsection{Task-aware Expert Loading}
\label{sec:slo_inf}
Running MoE-based models with a lower number of experts loaded into the GPU memory than the model originally trained can degrade the quality of the model output. However, our Observation~\ref{obs:slo_acc} indicates that certain tasks show greater sensitivity to token-to-expert routing accuracy while some tasks show little sensitivity to this accuracy. \sm{In the context of expert loading for a particular layer, we denote a task as either sensitive or insensitive.} \name has task-aware expert loading, which only considers the predicted experts of the sensitive tasks during the expert loading to reduce the expert loading latency while incurring minimal compromise on the accuracy of the model output.


\name runs its task-aware expert loader each time one or more new input requests are scheduled. As detailed in \S~\ref{sec:prediction}, the expert prediction module outputs the relative frequency with which tokens in a task type are likely to be routed to an expert. \name leverages this task type-based expert prediction to prioritize tasks that are more sensitive to token-to-expert routing accuracy. We define a task type's sensitivity to routing accuracy on an MoE layer basis. This sensitivity is evaluated through offline profiling, where, for each task type, we progressively apply random token-to-expert routing, starting from the MoE layer closest to the input. We then record the accuracy drop as we progressively use less accurate routing. The details are in \S~\ref{sec:task-moe-sen}.

\begin{wrapfigure}[16]{c}{4cm}
    \centering
    \includegraphics[width=4cm]{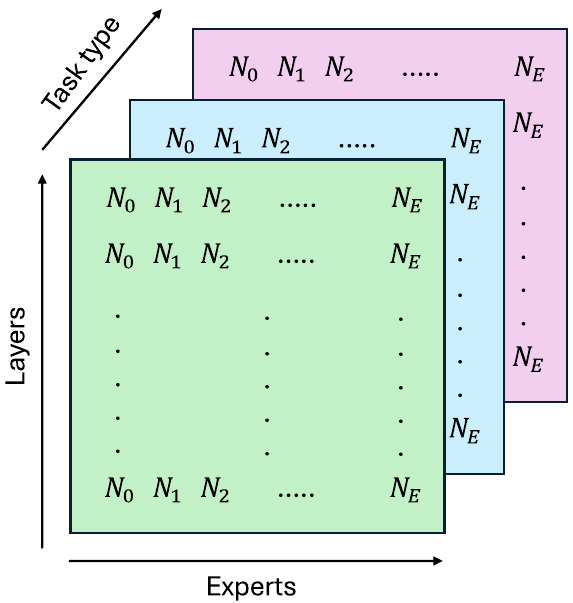} 
    \vspace{-0.2in}
    \caption{\name maintains the expected number of tokens per expert for each task type.}
   \label{fig:task_exp_tokens}    
\end{wrapfigure}

The MoE layers with inaccurate routing that yield output with accuracy higher than a threshold (e.g., 85\%) are marked as insensitive to the particular task.
In each iteration of expert loading, \name first computes the expected number of tokens per expert per task type and assembles the computed data in a 3D array as shown in Figure~\ref{fig:task_exp_tokens}.
\sm{We obtain the expected number of tokens per expert with an objective to maximize the number of correct token-to-expert routing. We hypothesize that maximizing the number of correctly routed tokens per expert will ensure the least impact on the accuracy. The rationale behind such hypothesis is that if we reduce the error in token-to-expert routing for a given number of experts, the accuracy drop will decrease. }


Let $N_i$ be the expected number of tokens that will be routed to expert $i$ in an arbitrary MoE layer for a given task type. Also, let $s\in\{0,1\}$ be a binary variable to indicate whether a token is sensitive to token-to-expert routing accuracy of the task type to the particular MoE layer, $f_i$ be predicted token-to-expert routing frequency, and $T$ be the number of requests belonging to the given task type \sm{that are already running}. We compute the value of $N_i$ using the following equation:\vspace{-0.1in}
\begin{eqnarray}
    N_i = (\sum_{j = 0}^{T-1} W_j + T \cdot W_o) \cdot s \cdot f_i
    \label{eqn:exp_tokens}\vspace{-0.3in}
\end{eqnarray}where $W_n$ denotes the number of input tokens in the $n$th request and $W_o$ denotes the expected number of generated tokens.\looseness=-1



\name utilizes Equation~\eqref{eqn:exp_tokens} to determine the expected number of tokens per expert for a given set of requests that are to be scheduled and the requests that are already running by the inference engine. \sm{\name obtains the total expected number of tokens per expert by summing $N_i$s belonging to all task types}. Next, \name sorts the experts at each MoE layer based on the expected number of tokens likely to be routed in descending order. Subsequently, \name picks the top $L$ experts for every layer, where $L$ is set based on the memory budget. In the MoE architecture where the number of experts across MoE layers is different, different $L$ values are chosen for each layer. The values of $L$ is set by the system user. Finally, \name identifies the experts that are already loaded into the inference engine. The identified experts are then set to be loaded into the inference engine in the next dispatching of the inference requests.

\subsection{Task-aware Request Scheduling}
\label{sec:sch_inf}
eMoE's request scheduler aims to minimize the impact on inference latency while ensuring compliance with the SLOs of running requests.
Under such setting, 
the existing scheduling systems (e.g.,~\cite{kwon2023efficient,dsfgen,agrawal2024taming,zhong2024distserve}) which 
neglect expert loading cannot optimally meet the latency SLO in \name. Furthermore the existing systems do not exploit the
Observations~\ref{obs:latency}, \ref{obs:slo_acc} which present some unique opportunities 
in terms of minimal impact on the inference latency. First, tasks' varying sensitivity to token-to-expert routing accuracy suggests that this phenomenon should be exploited to limit expert loading latency. 
Second, tasks have inherent latency requirements, leading to varying computational demands and differing impacts on the latency of running requests.
This is because scheduling a request may significantly increase the latency of the existing running requests due to the expert loading and a large number of input tokens. Therefore, it will be beneficial to delay the incoming request if the request has relaxed latency SLO requirement and the existing running requests are likely to complete the token generation in short period of time.
eMoE's request scheduler jointly considers user imposed latency SLOs, task-specific profiled generation length, and task-specific profiled sensitivity to token-to-expert routing accuracy to minimize the end-to-end inference latency.

Scheduling a new batch of requests introduces an initial upfront latency cost due to a large increase of the number of tokens per MoE layer in the model. This increase in the number of tokens equals to the number of input tokens. The increase in input size results in higher computation latency and increased all-to-all communication among the experts in MoE layer and thereby increasing the latency of the running requests.

Let $W$ be the number of input tokens in a new request to be scheduled, $G_i$ be the number of tokens to be generated by the $i$th request, $\Delta E$ be the expert loading latency due to the scheduling of new requests and $c$ be the average expert computation and communication latency for a single input token. We use task-specific profiling to obtain $G_i$. When a request is scheduled, $G_i$ is set to the average number of tokens generated by the task the request belongs to. After each iteration of token generation, $G_i$ is decremented. If $G_i$ reaches 0 but the request is not completed, $G_i$ is reset to a value proportional (e.g., 5\%) to its initial value. We also use a profiled value for $\Delta E$ and $c$. Also, let $n_i$ be the number of existing running requests that will complete generation after the $i$th request.  
We can express the expected latency of the $i$th request $t_i$ due to scheduling of a new request as follows:\vspace{-0.1in}
\begin{eqnarray}
    t_i = \Delta E + (W + n_i \cdot G_i )\cdot c + r_i 
    \label{eqn:com_lat}\vspace{-0.2in}
\end{eqnarray}
where $r_i$ denotes the run-time of request $i$ so far. For a new request, $r_i$ is set to zero and incremented as it runs on the inference engine.

\begin{algorithm}
\small
    \SetKwComment{tcp}{//}{}
    \caption{\name Request Scheduling Algorithm}
    \label{algo:scheduler}
    \KwIn{Waiting queue $Q_w$, Scheduled queue $Q_s$, Maximum number of tokens $T_{max}$}
    
    $T \gets Q_s.\text{length}$ \;
    $Q_w \gets$ Sort $Q_w$ by SLO stringiness 
    
    \For{$R \in Q_w$}{
        \If{$R.\text{inputTokens} + T < T_{max}$}{
            \If{$S.\text{NewExpectedLatency} < S.\text{SLO}, \forall S \in Q_s : S.\text{ExpectedLatency} < S.\text{SLO}$}{
                $Q_s \gets Q_s \cup \{S\}$ \;
                $T \gets T + R.\text{inputTokens}$ 
            }
        }
    }
    
    \KwRet $Q_s$ 
\end{algorithm}

\name adopts a greedy approach in its iterative scheduling algorithm. Algorithm~\ref{algo:scheduler} shows the pseudocode of eMoE's scheduling algorithm. At each iteration, \name picks the request with the most stringent latency SLO and attempts to schedule the request (\emph{line 2-6}). We measure a request's SLO stringiness by how quickly the first response token will be generated. The algorithm first checks if scheduling the request will exceed the total number of tokens that the inference engine can process. If the total number of tokens exceeds the limit, the request is kept in the waiting queue (\emph{line 2}). Otherwise, the request is scheduled if it will not lead to SLO violations of the requests already scheduled (\emph{line 3}). The request scheduler runs whenever a new request arrives or a running request completes token generation.


\section{Implementation}
We implemented \name using the Python programming language. The expert prediction model is implemented using Pytorch~\cite{NEURIPS2019_bdbca288}. The expert prediction model runs on a separate process besides the inference engine. We utilize DeepSpeed-FastGen~\cite{dsfgen} as the inference engine. The inference engine wraps Huggingface~\cite{wolf-etal-2020-transformers} models for inference serving. 

We extended the Huggingface model codes to augment functionality for expert loading. Specifically, we maintained Python multiprocessing lock for each MoE layer and wrapped the MoE layer's computation inside the lock. The lock is used by the expert loader process to synchronize the expert loading operation with the MoE layer computation. Furthermore, We wrapped the expert loading of an MoE layer with a CUDA event, which the corresponding MoE layer synchronizes with.
This prevents an MoE layer from using stale model weights. 

\name asynchronously loads experts in an MoE layer form host to device via $torch.Tensor.copy\_(non\_blocking=True)$ to overlap the loading process with the computation by the non-expert layers. We condition the loading of experts in an MoE layer contingent upon the completion of the loading process from the previous layer. We made such design decision to prevent the saturation of the PCIe channel bandwidth. 

\section{Performance Evaluation}
In this section, we first evaluate eMoE's end-to-end latency performance in \S \ref{e2e} which demonstrate the effectiveness of \name. We evaluate \name's impact on accuracy in \S \ref{eva-acc}. Then we demonstrate the performance  of expert predictor in \S \ref{sec:predictor_performance}. We then evaluate eMoE's inference performance with expert predictor (\S \ref{performance_prediction}) and show the impacts of reducing memory consumption (\S \ref{effect_loading}). Furthermore, we evaluate the sensitivity of \name on different parameters \S \ref{sec:task_acc_sen}. We present the time overhead of our methods in \S \ref{sec:ex_overhead}.
Unless otherwise specified, we use the same experiment setting as detailed in \S \ref{sec:analysis_sett}. 
Our key results include:
\squishlist

  \item eMoE reduces inference latency by up to 17\%. 
  
  \item eMoE reduces memory consumption up to 80\% while maintaining accuracy. By optimizing memory usage, it enables processing prompts 40$\times$ longer, batches 4.5$\times$ larger, and achieves 1.5$\times$  higher throughput (tokens/second).

 


    

\squishend

\subsection{Evaluation on End-to-End Inference Latency}
\label{e2e}
\begin{figure*}[htb]
    \centering
    \includegraphics[width=17cm, height=3.5cm]{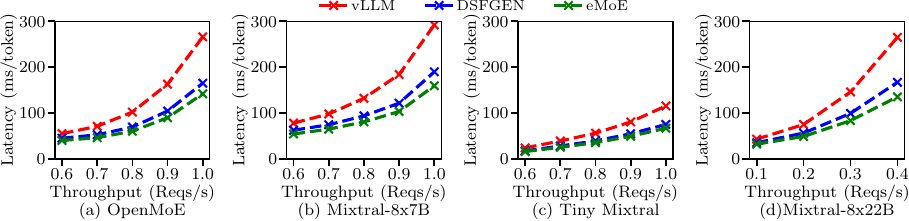}
    \vspace{-0.1in}
    \caption{End-to-End latency performance evaluations.}
    \vspace{-0.1in}
    \label{fig:lat_vs_thr}
\end{figure*}

Figure~\ref{fig:lat_vs_thr} shows the end-to-end inference latency of \name, vLLM, and DeepspeedFastGen for OpenMoE, Mixtral-8x7, TinyMixtral, and Mixtral-8x22B models. We compared with these two methods because they outperform other approaches. In the figure, the X-axis denotes the arrival rate of request per second and the Y-axis denotes the average inference latency. In the experiment, we varied the throughput in a step 0.1 request/second and generate tokens until the end of sequence (EOS) token. We only show the results for higher throughput for the interest of clarity. For Mixtral-8x22B model, the results include throughput upto 0.4, which is the maximum throughput supported by our experiment setup.

From the figure, we see that for all four models, \name outperforms DeepspeedFastGen and vLLM. Among the compared methods, vLLM performs the worst. DeepspeedFastGen's superior performance over vLLM is due to its kernel level optimizations tailored to MoE based models. \name also achieves superior performance compared to vLLM as \name is implemented on top of DeepspeedFastGen. 

From the figure we see that, \name consistently outperforms DeepspeedFastGen for a given throughput. The highest improvement (17\%) comes with the Mixtral-8x22B model which is the largest among the four in terms of the number of parameters. On the contrary, the smallest improvement (9\%) comes from TinyMixtral model which is the smallest among the four. We see that with \name performance improvement is larger with the larger model. This is due to larger model being compute heavy and memory heavy. \name runs with fewer experts which allows it to reduce the compute latency and global GPU memory read operations. Therefore, with larger model \name performs better.

\subsection{Evaluation on Inference Accuracy}
\label{eva-acc}
\begin{figure}[!t]
    \centering
    \includegraphics[width=8.5cm, height=3.5cm]{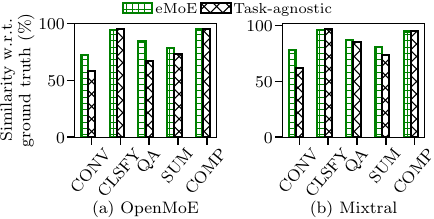}
    \caption{Accuracy with different tasks.}
    \label{fig:exp_acc_slo}
\end{figure}

Figure~\ref{fig:exp_acc_slo} shows the average accuracy achieved by the proposed task-aware expert loading and the task-agnostic expert  loading methods. In the task-agnostic loading methods, the expert prediction model's output for all the tasks in Table~\ref{table:data_tasks} was utilized in the expert loading. 
In the figure, the X-axis denotes different tasks and Y-axis 
represents the similarity of the model's output with the partially loaded expert to the model's output to the fully loaded experts. 
We refer the output of the latter as the ground truth. To measure the similarity with respect to ground truth, we pass the two outputs to a pretrained BERT~\cite{devlin2018bert} to measure the similarity. 
From the figure, we see that for all the tasks, \name performs better if not the same as the task-agnostic method. We see that for text classification and text similarity measure tasks, both methods perform almost the same. Recall that these two tasks are less sensitive to the accuracy of expert routing as indicated in \S \ref{sec:task-moe-sen} . However, for conversation, question-answer, and text summarizing tasks, we see that \name achieves better accuracy than the task-agnostic method.

\subsection{Performance of Expert Predictor} 
\label{sec:predictor_performance}
To find a suitable model for expert prediction, we conducted experiments on different variants of BERT \cite{devlin2018bert} and GPT-2 \cite{radford2019language} models from Huggingface \cite{wolf-etal-2020-transformers,wolf2020huggingfaces} and compare their results. We collected the expert routing data from the OpenMoE and Mixtral models, and use 80\% of the data for training and 30\% for testing. 
The results are shown in Table \ref{table:predictor}. The precision, recall and F1-score for Mixtral are  0.712, 0.713, 0.705 and for OpenMoE are 0.696,0.689, 0.680.
We observe that BERT-XLNet performs better than other models. 
XLNet \cite{yang2019xlnet} learns bidirectional contexts by maximizing the likelihood over all possible permutations of the input sequence, capturing more complex relationships between elements in the sequence, which leads to higher accuracy than other models. Therefore, we chose XLNet for our system. 


\vspace{-0.1in}
\begin{table}[!htb]
\centering
\scriptsize
\caption{Performance of the expert predictor.}
\begin{tabular}{p{1.5cm}p{0.8cm}p{0.8cm}p{0.8cm}p{0.8cm}}
\hline
Models & \# of  & Accuracy & Accuracy & Routing  \\

 & parameters & eMoE-L & eMoE-A & data \\
\hline
GPT-2 & 0.115B & 50\% & 49\% & OpenMoE \\
\hline
BERT-base & 0.108B & 13\% & 12\% &OpenMoE \\
\hline
BERT-XLNet & 0.108B & 70\%  & 69\% & OpenMoE \\
\hline
GPT-2 & 0.115B & 51\% & 51\% & Mixtral \\
\hline
BERT-base & 0.108B & 16\% & 16\% & Mixtral \\
\hline
BERT-XLNet & 0.108B & 71\% &71\% & Mixtral \\
\hline
\end{tabular}
\label{table:predictor} 
\end{table}
\vspace{-0.05in}

\subsection{Inference Performance with Expert Predictor}
\label{performance_prediction}
We compared eMoE-L and eMoE-A (with only the expert predictor invoked) against Baseline, Random \cite{zuo2021taming}, Pre-Gated MoE \cite{hwang2024pre}, and MoEInfinity \cite{xue2024moe}. The Baseline method loads all experts into GPU memory without using an expert predictor. The Random \cite{zuo2021taming} method randomly selects experts for tokens. 
The details of Pre-GatedMoE and MoEInfinity are explained in  \S\ref{sec:existing}. We also tested a variant of eMoE, in which the expert predictor (eMoE-A) is called for every prompt, denoted as eMoE-E. 

\begin{figure*}[htb]
\centering
{\includegraphics[width=0.96\textwidth]{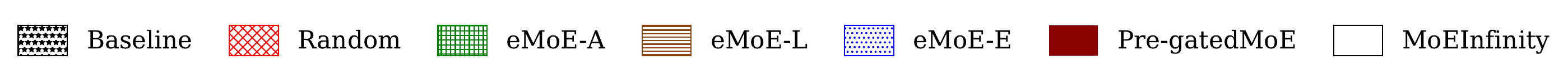}}
\vspace{-0.25in}
\end{figure*}

\begin{figure*}[htb]
	
	\centering
	\subfloat[OpenMoE.\label{fig:acc_om}]   {\includegraphics[height=1.9cm,width=4.2cm]{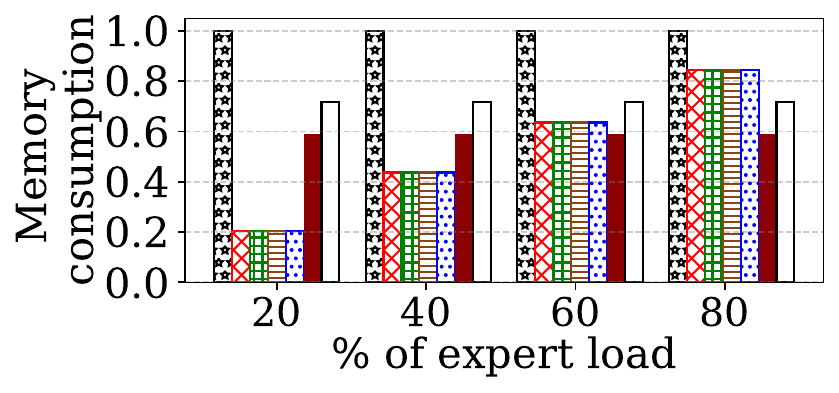}}
 \subfloat[TinyMixtral.\label{fig:acc_mx}]{\includegraphics[height=1.9cm,width=4.3cm]{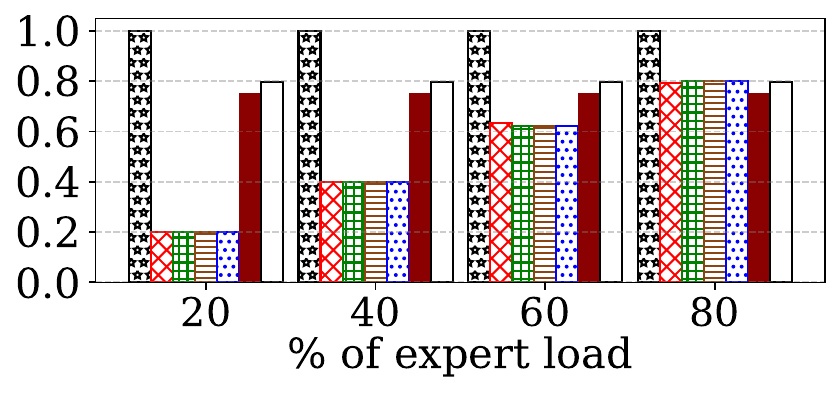}}
  \subfloat[Mixtral-8x7B.\label{fig:acc_mx}]{\includegraphics[height=1.9cm,width=4.3cm]{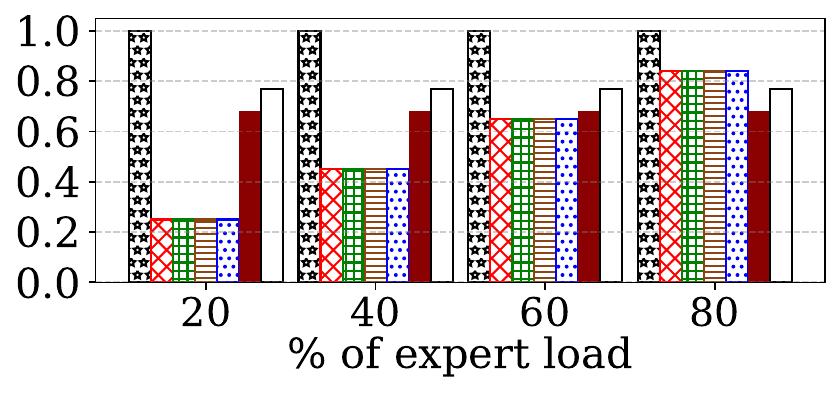}}
   \subfloat[Mixtral-8x22B.\label{fig:acc_mx}]{\includegraphics[height=1.9cm,width=4.3cm]{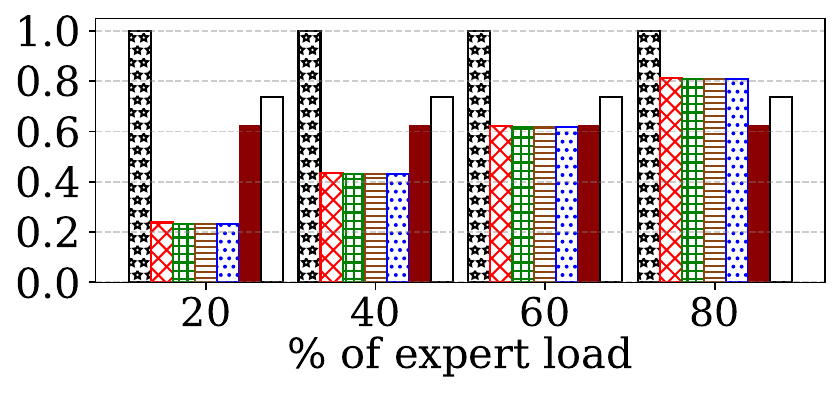}}
   \vspace{-0.1in}
	\caption{Memory consumption.}
 \label{fig:mem_ep_vary}
	\vspace{-.1in}
\end{figure*}

\begin{figure*}[!t]
	\vspace{-0.05in}
	\centering
	\subfloat[OpenMoE.\label{fig:acc_om}]   {\includegraphics[height=2.2cm,width=4.2cm]{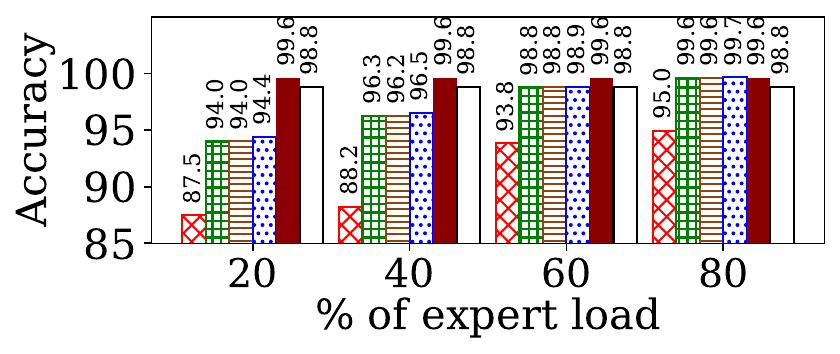}}
 \subfloat[TinyMixtral.\label{fig:acc_mx}]{\includegraphics[height=2.2cm,width=4.2cm]{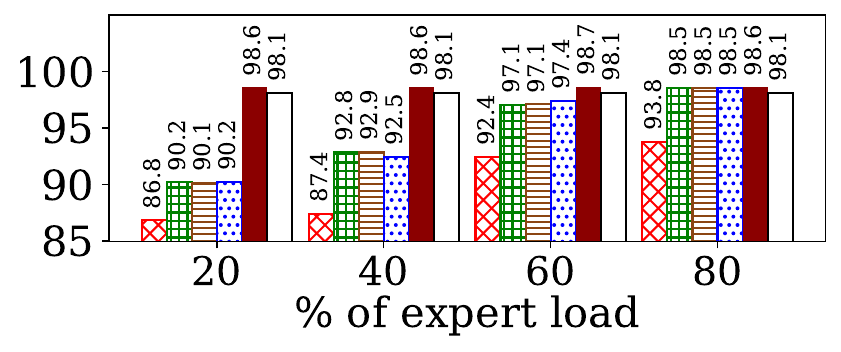}}
  \subfloat[Mixtral-8x7B.\label{fig:acc_mx}]{\includegraphics[height=2.2cm,width=4.3cm]{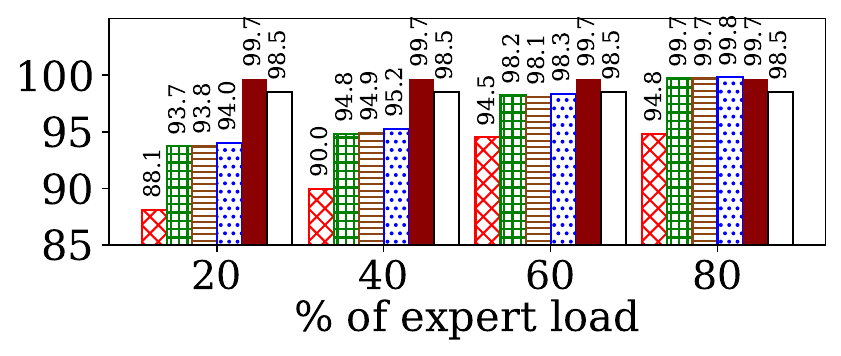}}
   \subfloat[Mixtral-8x22B.\label{fig:acc_mx}]{\includegraphics[height=2.2cm,width=4.3cm]{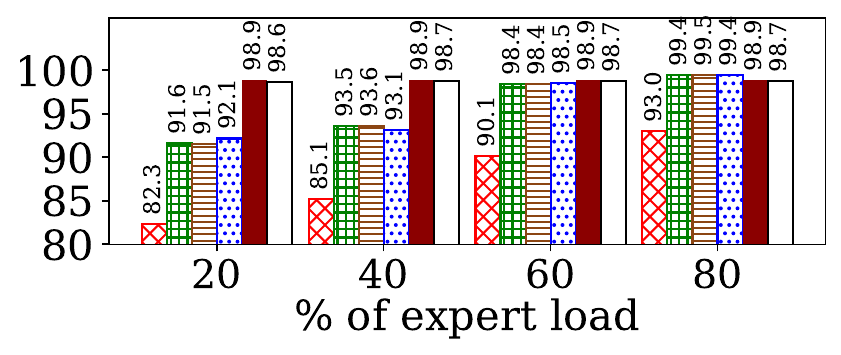}}
   \vspace{-0.1in}
	\caption{Accuracy.}
    \vspace{-0.1in}
 \label{fig:acc_ep_vary}
	
\end{figure*}

\begin{figure*}[!t]
	\vspace{-0.05in}
	\centering
	\subfloat[OpenMoE.\label{fig:acc_om}]   {\includegraphics[height=2.1cm,width=4.2cm]{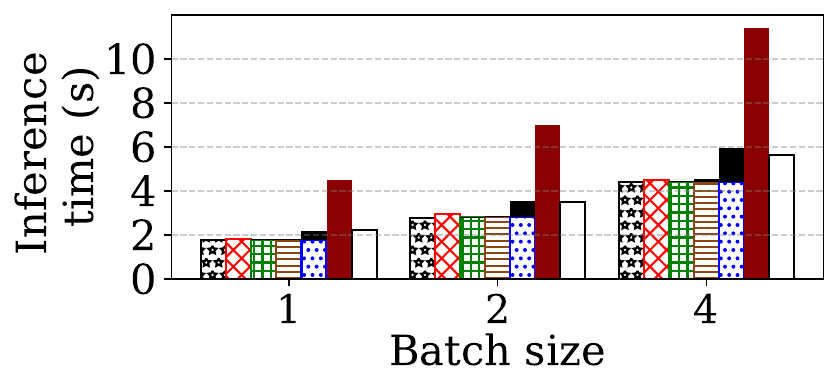}}
 \subfloat[TinyMixtral.\label{fig:acc_mx}]{\includegraphics[height=2.1cm,width=4.3cm]{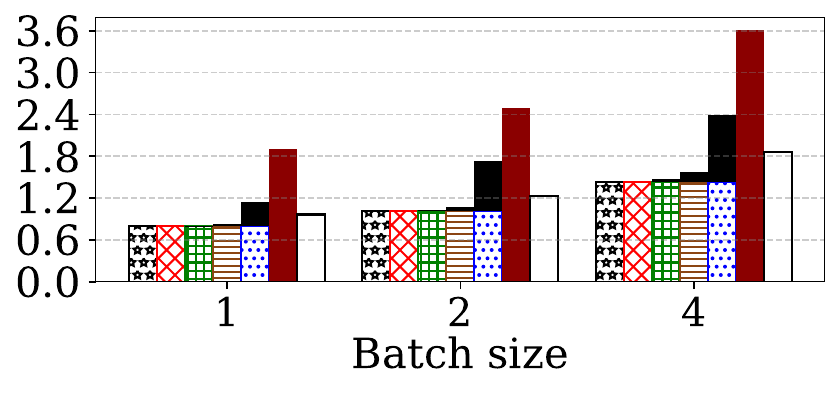}}
  \subfloat[Mixtral-8x7B.\label{fig:acc_mx}]{\includegraphics[height=2.1cm,width=4.3cm]{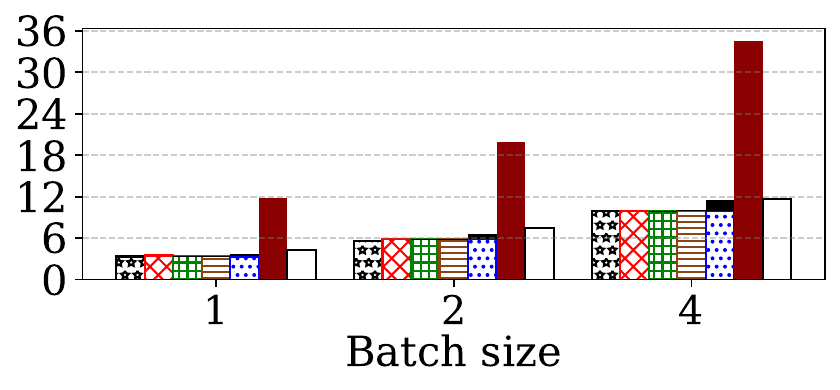}}
   \subfloat[Mixtral-8x22B.\label{fig:acc_mx}]{\includegraphics[height=2.1cm,width=4.3cm]{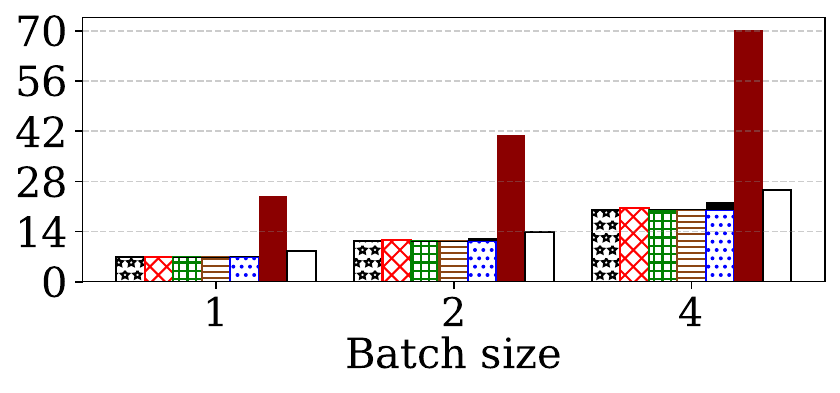}}
   \vspace{-0.1in}
	\caption{Inference time.}
    \vspace{-0.15in}
 \label{fig:time}
	
\end{figure*}


\textbf{Memory Consumption and Accuracy.} 
Figures 14 and 15 illustrate memory consumption and accuracy for various percentages of experts loaded into GPU memory. Baseline, which loads all experts, results in the highest memory usage. eMoE variants reduce memory consumption by 20\%-80\% while maintaining 93.7\%-99.8\% accuracy. With 60\% of experts loaded, eMoE achieves 98.2\%-98.8\% accuracy, and with 80\%, it reaches 99.6\%-99.7\%. This high accuracy stems from eMoE's predictive approach, which minimizes incorrect expert selection. In contrast, the Random method shows significantly lower accuracy, underperforming eMoE by 4.7\%-8.6\% for OpenMoE and 4.1\%-6.5\% for Mixtral.
Pre-gatedMoE (98.6\%-99.7\%) and MoEInfinity (98.1\%-98.7\%) show comparable accuracy to eMoE, as they employ similar expert prefetching strategies. Among eMoE variants, eMoE-E achieves slightly higher accuracy (0.1\%-0.4\%) than eMoE-A and eMoE-L by calling the predictor for each prompt, improving expert selection accuracy.

\textbf{Inference Time.} 
Figure \ref{fig:time} shows the average inference time for different batch sizes and the overhead for calling the expert predictor (upper black part). A detailed analysis of time overhead is in \S \ref{sec:ex_overhead}. We can observe that the inference time increases with batch size due to the greater computational workload. Both eMoE-A and eMoE-L perform similarly to the Baseline and Random methods because they only invoke the expert predictor periodically. In contrast, Pre-GatedMoE demonstrates a 2.4x-3.5x higher inference time compared to eMoE variants. Pre-GatedMoE prefetches experts to overlap data transfer with computation at the current layer. However, expert transfers between the CPU and GPU cause memory and bandwidth contention, slowing data transfer and processing, especially when the GPU is simultaneously handling computations, leading to increased latency. MoEInfinity shows 1.25x-1.5x higher inference time compared to eMoE. MoEInfinity employs prefetching for the later layers and uses stored experts for the initial layers. Since it also employs prefetching during inference, its inference time is longer than that of eMoE.
However, eMoE-E has a 9\%-34\% higher overhead as it calls the predictor for every prompt, resulting in longer inference time. Although eMoE-E achieves slightly higher accuracy, the trade-off between accuracy and increased inference time makes eMoE-A and eMoE-L more practical choices than eMoE-E for our system.

\subsection{Impacts of Reducing Memory Consumption}
\label{effect_loading}

\begin{figure}[htb]
\vspace{-0.05in}
		\centering
	\subfloat[OpenMoE.\label{fig:seq_om}]   {\includegraphics[width=4.3cm,height=2.4cm]{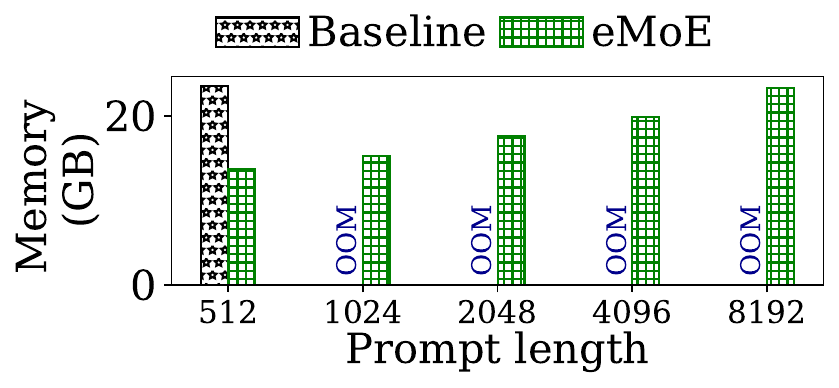}}
 \subfloat[Mixtral-8x7B. \label{fig:seq_mx}]{\includegraphics[width=4.3cm,height=2.4cm]{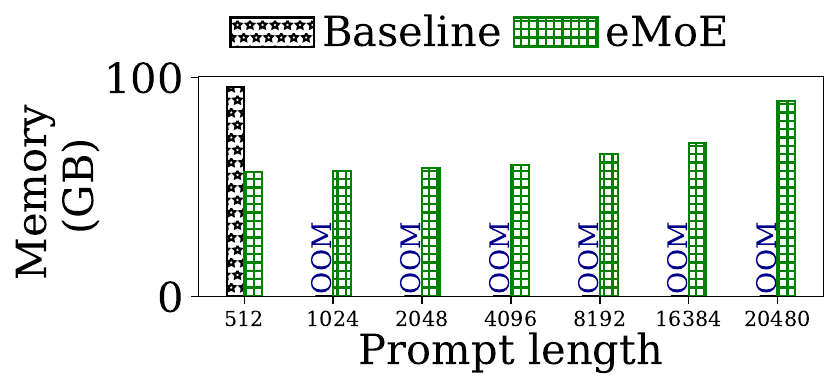}}
 \vspace{-0.1in}
	\caption{Memory consumption vs. prompt length.}
    \vspace{-0.05in}
 \label{fig:seq_ep}
\end{figure}

\textbf{Impact on Prompt Length.}
eMoE reduces GPU memory consumption, allowing it to process longer prompts. We tested eMoE-A and eMoE-L with 60\% of experts loaded and gradually increased the prompt length until memory consumption matched the Baseline. Figure \ref{fig:seq_ep} shows memory usage for different prompt lengths. At a length of 512, the Baseline OpenMoE uses 24GB, while eMoE-A and eMoE-L use 14GB. At a length of 8192, eMoE's memory consumption matches the Baseline, allowing it to handle prompts 16x longer. Similarly, the Baseline Mixtral uses 96GB for a prompt of 512, while eMoE-A and eMoE-L use 59GB, enabling them to process prompts up to 40x longer.

\begin{figure}[htb]
\vspace{-0.15in}
		\centering
        {\includegraphics[width=8.5cm, height=2.8cm]{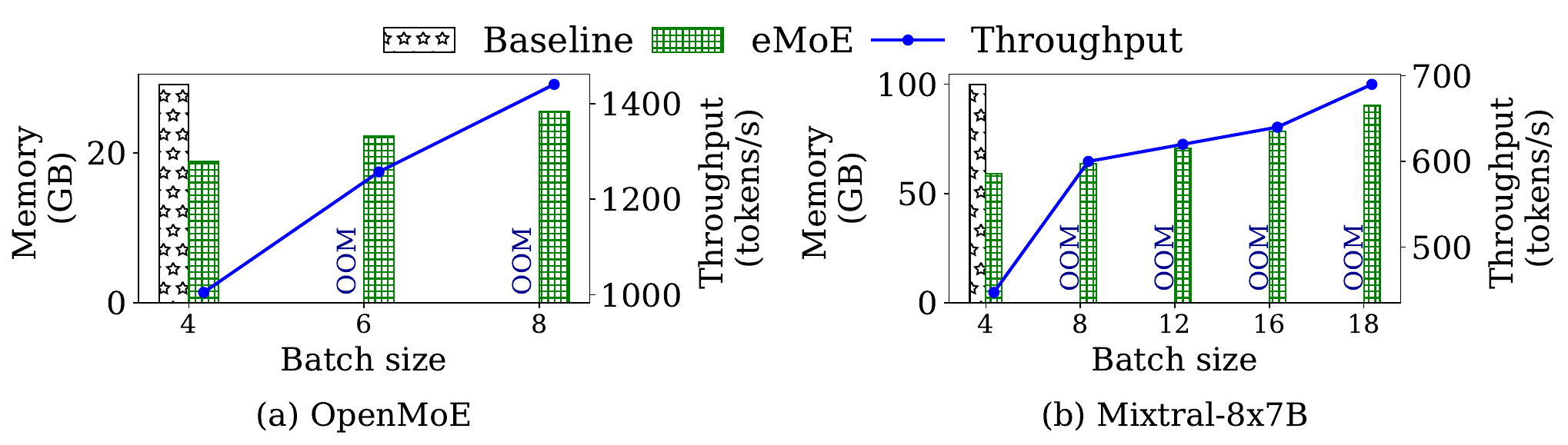}}
 \vspace{-0.2in}
 \caption{Memory consumption vs. batch size.}
 \label{fig:batch_ep}
 \vspace{-0.2in}
\end{figure}

\textbf{Impact on Batch Size and Throughput.} 
We conducted experiments to determine the maximum batch size eMoE-L and eMoE-A can handle with the saved memory. We progressively increased the batch size until its memory consumption matches Baseline's memory consumption.
Figure \ref{fig:batch_ep} shows memory consumption for varying batch sizes. The Baseline OpenMoE uses around 29GB for a batch size of 4, while eMoE-L and eMoE-A use about 19GB. As the batch size increases to 8, the memory consumption of eMoE-L and eMoE-A rises to 29GB, indicating that they can efficiently process up to 8 batches, handling 2x more batches than Baseline.
Similarly, the Baseline Mixtral consumes about 100GB for a batch size of 4, while eMoE-L and eMoE-A use approximately 59GB. For a batch size of 18, their memory consumption matches 100GB.
This demonstrates that eMoE-L and eMoE-A can efficiently handle up to 18 batches, processing 4.5x more batches than the Baseline model.
We also measured throughput, which refers to the number of tokens processed per second at increased batch sizes. The blue lines in Figure~\ref{fig:batch_ep} show eMoE's throughput increase. OpenMoE achieves 1.3x to 1.4x improvement as batch size grows from 4 to 8, while Mixtral sees a 1.3x to 1.5x boost from 4 to 18. These results show that eMoE can process up to 1.5x more tokens per second compared to the Baseline, demonstrating both memory efficiency and improved processing speed.

\begin{table}[!htb]
\centering
\scriptsize
\caption{Performance comparison with Quantized model}
\begin{tabular}{c|c|c}
\hline
Methods & Memory (GB) & Accuracy   \\
\hline
eMoE & 64.504 & 98.2\% \\
\hline
Quantized-8bit & 57.814  & 95.2\% \\
\hline
Quantized-4bit & 30.487 & 91.5\% \\
\hline
\end{tabular}
\label{table:quantized} 
\end{table}
\textbf{Benefits over Quantized models.}
We conducted experiments to evaluate the benefits of using eMoE compared to a quantized model. 
To assess eMoE's performance, we compared it to the 4-bit and 8-bit quantized versions of the Mixtral 8x7B model, recording both memory usage and accuracy. As shown in Table~\ref{table:quantized}, the 4-bit model reduced memory by 2.2x and the 8-bit model by 1.2x compared to eMoE-A, but at the cost of a 3.06\%-6.82\% accuracy drop. These results demonstrate that eMoE offers a better balance of memory efficiency and accuracy, outperforming quantized models.

\subsection{Sensitivity Analysis}
\label{sec:task_acc_sen}
\subsubsection{Sensitivity To Task Classification Accuracy}
\begin{wrapfigure}[16]{c}{4cm}
    \centering
    \includegraphics[width=4cm]{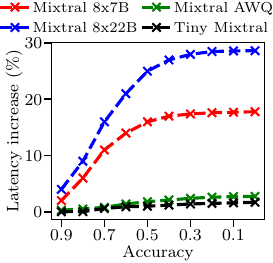} 
    \vspace{-0.2in}
    \caption{Impact on latency under different task classification accuracy.} \vspace{-0.3in}
   \label{fig:task_sen_acc}    
\end{wrapfigure}

Figure~\ref{fig:task_sen_acc} shows the percentage change in latency with respect to the latency achieved with 100\% task classification accuracy. In the experiment, we randomly introduced an inaccuracy in task label for a given task accuracy, record the average inference latency, and computed the percentage change in latency with respect to the ground truth case. 
From the figure, we see that with the increase in inaccuracy in task identification, the inference latency increases for all four models. The increase in latency mainly stems from excess host-to-device expert loading. Note that \name uses its task-aware expert loading method to selectively load experts based on expert sensitivity. \name relies on accurate task type identification to achieve this objective. However, with more inaccuracy, \name starts to load the experts from the host when it is not required to. However, even with 20\% inaccuracy, the impact on the latency is less than 10\% even for the largest model. When the inaccuracy becomes excessively high, \name loads the experts from the host most of the times, which causes the increase in latency to plateau.

\subsection{Time Overhead}
\label{sec:ex_overhead}
We tested the time overhead of calling the expert predictor for eMoE-A and eMoE-L. We measured the time taken for expert prediction and moving the experts to GPU, then compute the average. The time for eMoE-A and eMoE-L averages at $\sim$0.381s and $\sim$1.387s for OpenMoE, and averages at $\sim$0.334s and $\sim$4.211s for Mixtral-8x7B. The expert predictor is called every 40 prompts, as determined empirically (Figure \ref{fig:analysis_same}). For OpenMoE, the time overhead for eMoE-A and eMoE-L is 0.47\% and 1.69\% of the average inference time per request. For Mixtral, it is 0.24\% and 3.11\%, respectively. Mixtral incurs higher time overhead due to its 32 layers, each requiring prediction and transfer time, whereas OpenMoE has only 4 layers. The task-aware request scheduler runs on the CPU, incurring no additional overhead.

\section{Related Work}\vspace{-0.05in}
\label{sec:rl}
\textbf{Mixture-of-Expert models.}
Recent research has focused on enhancing large-scale models using MoE-based architectures \cite{shazeer2017outrageously, zuo2021taming, li2020video, nie2022hetumoe, qin2020multitask, he2023merging, dai2022stablemoe, chowdhury2023patch, pmlr-v202-zhou23c, lewis2021base, gao2022parameter, wang2022adamix,dai2024deepseekmoe, kudugunta2021beyond, kang2024self, zhao2024hypermoe, zhang2022samoe}. Google's Switch Transformer \cite{fedus2022switch} pioneered MoE exploration, dynamically selecting different experts for different parts of the input sequence. GShard \cite{lepikhin2020gshard} scales MoE to a trillion parameters and uses expert parallelism. Other models such as PaLM \cite{JMLR:v24:22-1144}, GaLM \cite{DBLP:journals/corr/abs-2112-06905} from Google, and M6-T \cite{DBLP:journals/corr/abs-2105-15082} from Alibaba have also excelled in language and multi-modal tasks. MoEfication \cite{zhang2021moefication} transforms conventional models into MoE models by splitting FFN parameters into multiple expert partitions. DeepSpeed Zero3 \cite{rajbhandari2020zero} collects parameters from other GPUs dynamically to support extremely large models.

\noindent\textbf{MoE Training and Inference Systems.}
A group of methods focus on improving the performance and scalability of MoE models \cite{shen2023semoe, rajbhandari2022deepspeed, liu2023janus, MLSYS2023_5616d34c, he2022fastermoe, he2021fastmoe, zhai2023smartmoe, li2023accelerating, MLSYS2022_37385144, MLSYS2023_5a54f793, 288582, shen2022se, shi2024schemoe, zhong2024distserve, agrawal2024taming, gale2023megablocks}. Deepspeed-MoE \cite{rajbhandari2022deepspeed} utilizes tensor slicing  and expert-slicing to split parameters across multiple GPUs and to leverage memory bandwidth efficiently. 
Janus \cite{liu2023janus} uses a data-centric paradigm that focuses on keeping data in place and moving experts between GPUs to reduce communication workload. 
Tutel \cite{MLSYS2023_5616d34c} 
uses adaptive parallelism switching and adaptive pipelining to manage the dynamic workloads of MoE. 
Faster-MoE \cite{he2022fastermoe} uses a shadow expert to address imbalanced workloads and a fine-grained scheduler for achieving asynchronous all-to-all communication. FastMoE \cite{he2021fastmoe} offers a hierarchical interface that enables flexible MoE model design and easy adaptation to various applications, including Transformer-XL \cite{dai2019transformer} and Megatron-LM \cite{shoeybi2019megatron}. SMARTMoE \cite{zhai2023smartmoe} is an automatic parallelization system for distributed training for MoE-based models. Lina \cite{li2023accelerating} prioritizes all-to-all over all-reduce using tensor partitioning to improve training step time. Pre-Gated MoE~\cite{hwang2024pre}, MoEInfifnity \cite{xue2024moe} and EdgeMoE \cite{yi2023edgemoe} leverage pipelining to overlap computation and the transfer of experts from CPU to GPU. MoE-Lightning~\cite{cao2024moe} also leverages CPU-GPU pipelining and a Hierarchical Roofline Model to achieve higher throughput. Compared to these methods, \name significantly reduces memory consumption while preserving accuracy and also minimizes end-to-end inference latency.





\vspace{-0.05in}
\section{Conclusion}\vspace{-0.1in}
We present \name, a memory-efficient MoE inference system that uses an expert prediction model to load experts and incorporate several advanced methods to reduce latency. 
It has periodic expert invocation to avoid frequent expert prediction and loading. Moreover, we demonstrate that the input sensitivity to token-to-expert routing accuracy varies among tasks, prompting us to develop a task-aware expert loading method that prioritizes tasks with higher sensitivity. Additionally, we introduce a task-aware request scheduling method that optimizes request scheduling based on task-specific token generation length, task-aware expert loading latency, and latency SLO. Our experiments demonstrate that \name significantly outperforms existing MoE inference systems in terms of memory consumption and inference time while incurring a minimal impact on accuracy.


\bibliographystyle{unsrt}
\bibliography{paper}

\end{document}